# Characterizing the Variability in Face Recognition Accuracy Relative to Race

Krishnapriya K. S[1], Kushal Vangara[1], Michael C. King[1], Vítor Albiero[2], Kevin Bowyer[2]

[1]Florida Institute of Technology, [2]University of Notre Dame

**Abstract**

*Many recent news headlines have labeled face recognition technology as "biased" or "racist". We report on a methodical investigation into differences in face recognition accuracy between African-American and Caucasian image cohorts of the MORPH dataset. We find that, for all four matchers considered, the impostor and the genuine distributions are statistically significantly different between cohorts. For a fixed decision threshold, the African-American image cohort has a higher false match rate and a lower false non-match rate. ROC curves compare verification rates at the same false match rate, but the different cohorts achieve the same false match rate at different thresholds. This means that ROC comparisons are not relevant to operational scenarios that use a fixed decision threshold. We show that, for the ResNet matcher, the two cohorts have approximately equal separation of impostor and genuine distributions. Using ICAO compliance as a standard of image quality, we find that the initial image cohorts have unequal rates of good quality images. The ICAO-compliant subsets of the original image cohorts show improved accuracy, with the main effect being to reducing the low-similarity tail of the genuine distributions.*

## 1. Introduction

There have been numerous reports touting vast improvements in accuracy of face recognition matchers on increasingly challenging images of faces collected in the wild. However, there have also been several news items in the past two years to decry face recognition technology as being "biased" or "racist". One example headline is "Facial Recognition Is Accurate, if You're a White Guy" [1]. Three spurs to this news stream have been reports and press releases from the Georgetown Law Center on Privacy and Technology [2], the ACLU [3] and the MIT Media Lab [4]. Additionally, as reviewed in the next section, there is an existing body of scholarly research reporting on experiments in which the accuracy of face recognition is found to vary across gender, age or race.

The goal of our work is to understand why inequalities in face recognition accuracy occur, and what might be done to mitigate them. This paper reports on experiments using four face matchers and a large face image dataset available to the research community [11,12], focusing on recognition accuracy for African-American and Caucasian image cohorts. Novel contributions of our work include (a) insights into how genuine and impostor score distributions differ between African-American and Caucasian image cohorts, (b) results showing that these differences can lead to receiver operator characteristic (ROC) curves showing worse *or better* accuracy for the African-American cohort, and (c) results suggesting that there is a difference in average image quality between cohorts, but that this does not account for the differences in the impostor and genuine distributions.

Section 2 summarizes elements of previous work. Section 3 describes the dataset and matchers used in this work. Section 4 presents an analysis of false match rate (FMR) and false non-match rate (FNMR) as a function of the matcher decision threshold. Section 5 describes differences in the impostor and genuine distributions between cohorts. Section 6 considers how differences in image quality between cohorts are related to differences in accuracy. Sections 7 and 8 present conclusions and discussion points.

## 2. Related Work

Observations of face recognition accuracy varying across demographic groups date at least to the 2002 Face Recognition Vendor Test (FRVT) [5]. FRVT 2002 results suggested that men are more accurately recognized than women, older persons are more accurately recognized than younger persons, and that the difference in accuracy between males and females decreases with age.

In a 2009 review of 25 works, Lui et al. [6] found "… near complete agreement … that older people are easier to recognize than younger", that "… there is no consistent gender effect" and that "… no clear conclusions can be drawn about whether one racial group is harder or easier to recognize". Note that the only two racial groups for which there was enough data for the authors to venture a conclusion were Caucasian and East Asian.

Beveridge et al. [7] reported that older persons are more accurately recognized than younger, that "a trend is emerging suggesting men are easier to recognize than women", and that "Algorithms responded differently to race, but generally all non-Caucasian races except Black were easier to verify".

Phillips et al. [8] found that, at low false accept rates required for most security applications, a "Western" fusion algorithm recognized Caucasian faces more accurately than East Asian faces, and that an "East Asian" fusion algorithm recognized East Asian faces more accurately than Caucasian faces. They conclude that the results "suggest a need to understand how the ethnic composition of a training set impacts ... algorithm performance".

Klare et al. [9] report on experiments with six matchers and a large dataset of mug-shot images from the Pinellas County Sheriff's Office. With respect to age, they report that "all three commercial algorithms had the lowest matching accuracy on subjects grouped in the ages 18 to 30". With respect to gender, "each of the three commercial face recognition algorithms performed significantly worse on the female". And with respect to race, "all three commercial face recognition algorithms achieved the lowest matching accuracy on the Black cohort".

El Khiyari and Wechsler [10] report results using a small subset of the MORPH [11] dataset, 362 subjects and 1448 images each for an African-American and a Caucasian cohort. Comparing a VGG-based and a COTS matcher, they find substantially better ROCs for the VGG matcher. For the VGG matcher, the Caucasian ROC is much better than the African-American ROC. For the COTS matcher, the two ROCs are similar. (There is a PCA dimensionality reduction of the VGG feature vector, trained on the same data that performance is reported for, which may have resulted in optimistic VGG results.)

In summary, a very small number of peer-reviewed publications have compared face recognition accuracy between African-American and Caucasian image cohorts [9,10] ([4] looks at gender prediction, not recognition). None have reported differences in accuracy at the level of the impostor and genuine distributions, reported an instance of the African-American cohort having a better ROC curve than the Caucasian cohort, or investigated the possibility of differences in image quality between cohorts.

## 3. Experimental Dataset and Matchers

The MORPH dataset was assembled in the mid-2000s from public records, to support face aging research [11] and is widely used in the research community. It is particularly appropriate for this research, as it has a substantial number of images of African-American subjects. It consists of mugshot images, of resolutions 200x240 and 400x480. Example MORPH images are shown in Figure 1. We curated a version of MORPH Album 2 for this work. We found the MORPH meta-data to be highly accurate, so that curation involved changes for less than 1% of the original images: 259 of 53,633 images were deleted due to not containing a face, 140 images were deleted due to being repeated instances of an image, meta-data was corrected on 92 images, and 3 images were deleted due to apparently incorrect meta-data deemed too ambiguous to correct. (As examples of meta-data correction, the original meta-data for images 092829_3M35 and 072771_1M44 in Figure 1 was

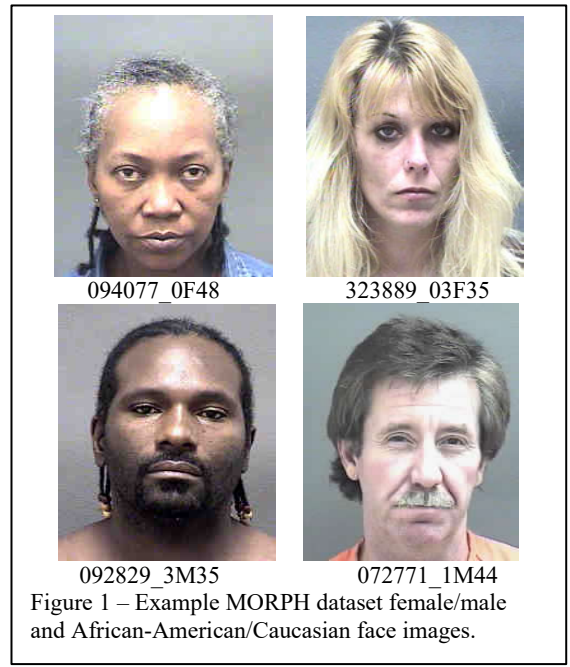

Figure 1 – Example MORPH dataset female/male and African-American/Caucasian face images.

Caucasian and African-American, respectively; these were corrected to African-American and Caucasian.) The curated dataset used in this work is 53,231 images of 13,119 subjects, split into an African-American cohort of 42,620 images of 10,350 subjects, and a Caucasian cohort of 10,611 images of 2,769 subjects.

We report results from four face matchers: commercial SDKs "COTS-A" and "COTS-B", and the popular CNN-based matchers VGG [13] and ResNet [14]. COTS-B is more recent than COTS-A; the matchers' names are not given due to license agreements. The VGG model used in our work is the one pre-trained using the VGGFace dataset [13]; features are taken from layer 16. The version of ResNet used is the one was pre-trained on the VGGFace 2 dataset [15]; features are taken from layer 50. No additional fine-tuning was done on the pre-trained weights for either VGG or ResNet. Cosine similarity metric was used to generate match scores for both.

For COTS-A, 26 of the 10,615 Caucasian face images (0.24%) and 109 of the 42,616 African-American face images (0.26%) had a failure-to-enroll (FTE) result. For COTS-B, 1 of the 10,615 Caucasian face images and 4 of the 42,616 African-American face images had a FTE. Only 1 of the 5 COTS-B FTE images was also an FTE with COTS-A. Factors that appear to contribute to FTE results include glasses, hair that covers parts of the face, closed eyes, and off-angle gaze. Due to the small number of FTE results and the variety of factors involved, the data is insufficient to support any conclusion about relative FTE rate for African-American and Caucasian.

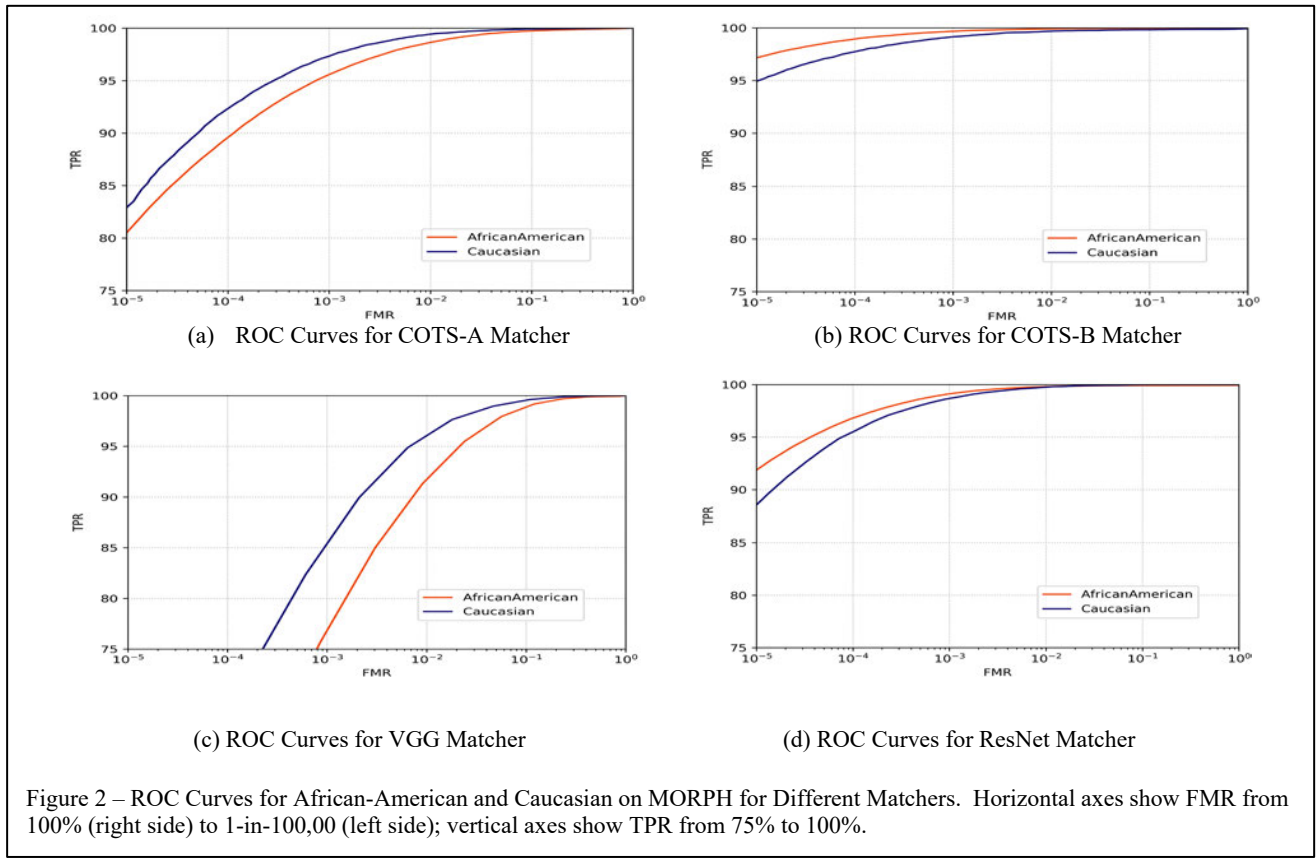

Figure 2 – ROC Curves for African-American and Caucasian on MORPH for Different Matchers. Horizontal axes show FMR from 100% (right side) to 1-in-100,00 (left side); vertical axes show TPR from 75% to 100%.

For VGG and ResNet, faces are detected using a pre-trained 68 landmark points shape predictor model [16] in the pre-processing stage. The region of interest around the detected face is cropped and resized to 224x224 pixel resolution for input to the CNN. With Dlib, 64 African-American and 14 Caucasian face images had a failure-to-detect result. In 4 cases, Dlib reported a second, noise face detection in a MORPH image. Those images were pre-processed manually.

## 4. FMR and FNMR Analysis

The FMR is a key statistic for characterizing the accuracy of a biometric system. Once an acceptable FMR range has been chosen for a given application, candidate matchers may be evaluated against a representative dataset to determine which has the lowest corresponding FNMR (or, highest verification rate). The ROC curve is a commonly used tool for performing a comparative analysis of competing algorithms on the same dataset as it provides a visual illustration of the tradeoff between FNMR and FMR. One can easily identify which matcher has the highest TPR at a specified FMR.

The ROC curves for the matchers are shown in Figure 2. We take particular note of TPRs at operationally relevant FMR values of $10^{-3}$ and where possible $10^{-4}$. The ROC curves for COTS-A and for VGG show better accuracy for Caucasians than for African-Americans. This is in agreement with the results of previous research [9,10]. However, the ROC curves for COTS-B and ResNet show better accuracy for African-Americans than for Caucasians. No previous publication has reported an instance of ROC results for an African-American image cohort showing better accuracy than the ROC for a Caucasian cohort. But the larger and more important point here is that ROC curves are generally not an appropriate way to compare face recognition accuracy across demographic cohorts. The typical use of ROC curves is to compare accuracy of different algorithms on the same dataset, and we are comparing accuracy of the same algorithm(s) across different datasets.

It is essential to note that a specified FMR is usually realized by different threshold values relative to the African-American and the Caucasian impostor distributions (See Section 5). The ROC curve is a plot of the verification rate, or 1-FNMR, as a function of FMR. Therefore, which cohort has the better ROC is determined by which has the better verification rate at the score threshold that realizes the specified FMR for that cohort.

Figure 3 shows FMR and FNMR curves. For all four matchers, the African-American FMR curve is higher than the Caucasian FMR curve. The size of the gap varies between matchers. For example, the gap is larger for the VGG matcher and smaller for the COTS-B matcher.

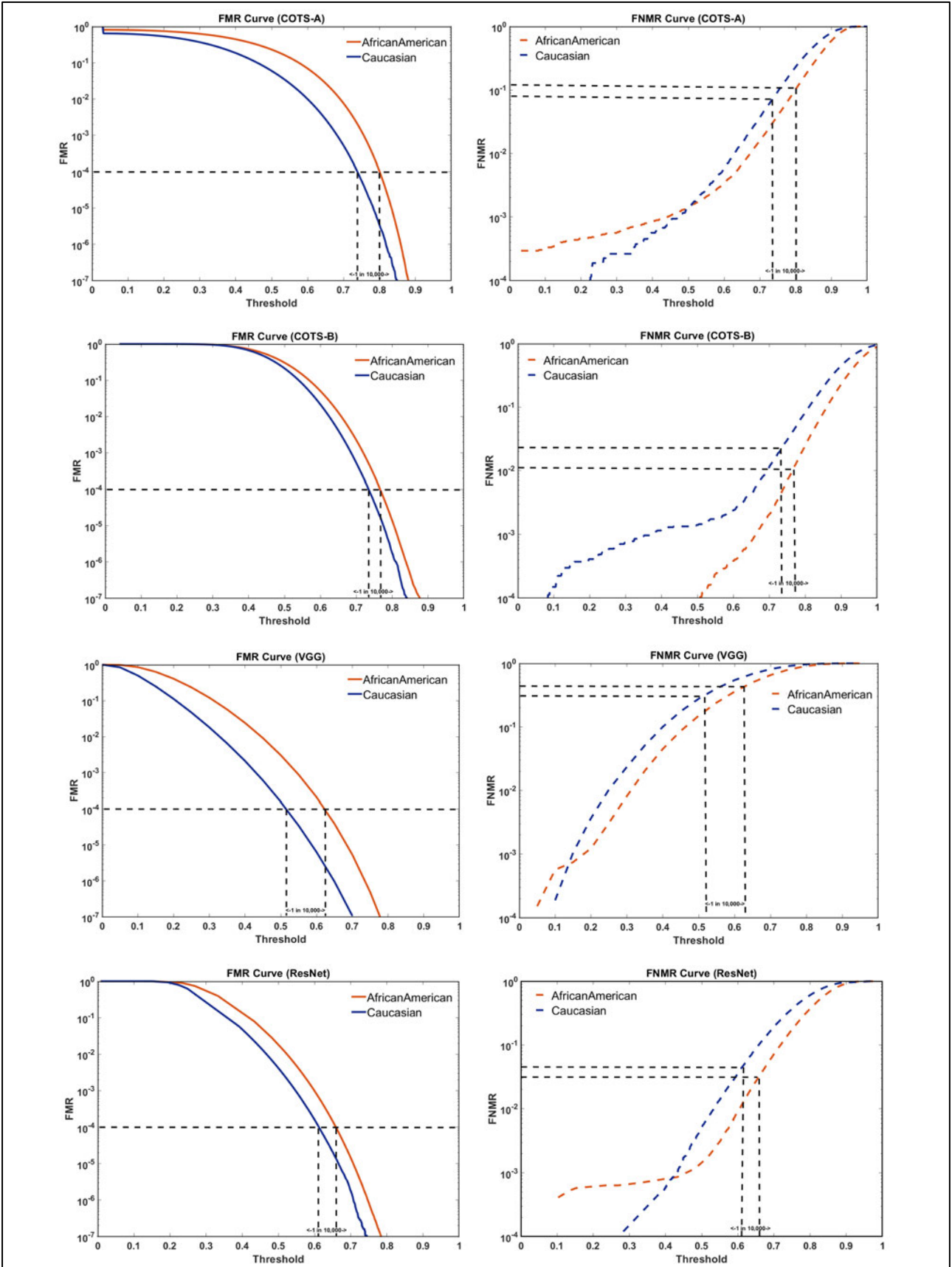

Figure 3 – FMR and FNMR Curves for Matchers on MORPH dataset.

A dashed line is drawn horizontally across each FMR plot at the 1-in-10,000 FMR level. The verification rate at a 1-in-10,000 FMR is often used for evaluating the performance of a face recognition system. The dashed line is followed down from where it intersects the two FMR curves, to indicate the threshold value that would realize the 1-in-10,000 FMR for each cohort. The larger the gap between the curves, the larger the disparity between the decision threshold values for 1-in-10,000 FMR. The range of operationally relevant decision threshold values is in the general range of the marked decision threshold values.

For the FNMR plots, the African-American curve is generally below the Caucasian curve. Again, the gap varies between matchers. For example, the gap is larger for COTS-B and smaller for COTS-A. For COTS-A, the curves even cross around a decision threshold value of 0.5. However, this is well outside of an operationally relevant range of decision threshold values. Also, referring back to the genuine distributions in Figure 2, it is clear that the FNMR values at a threshold of 0.5 are based on scarce data, and so the crossing is noisy and uncertain.

For each matcher, the decision thresholds found to realize a 1-in-10,000 FMR for that matcher are marked on the FNMR plots with a vertical dashed line up to intersect the FNMR curve and then horizontally over to the FNMR value. The difference in the FNMR for the two cohorts is a result of the difference in the thresholds that realize the 1-in-10,000 FMR for each cohort and the gap between the FNMR curves in that region. Following the dashed lines on the COTS-A and VGG FNMR curves shows that, at the threshold values where each cohort realizes a 1-in-10,000 FMR, the Caucasian cohort has a lower FNMR. However, following the dashed lines on the COTS-B and ResNet FNMR curves shows that, at the threshold values where each cohort realizes a 1-in-10,000 FMR, the African-American cohort has a lower FNMR.

Tracing through the different thresholds that realize a 1-in-10,000 FMR for each cohort, and how those translate to different FNMR values, emphasizes two important findings. One, it is clear that with the African-American image cohort having a generally higher FMR and a generally lower FNMR, the ROC curve for the African-American cohort could in principle be worse or better than the ROC for the Caucasian cohort. Two, *an ROC-level comparison between cohorts can show either cohort as having better accuracy when in fact there is a consistent difference in the underlying FMR and FNMR*. In effect, the ROC-level analysis can "hide" important information.

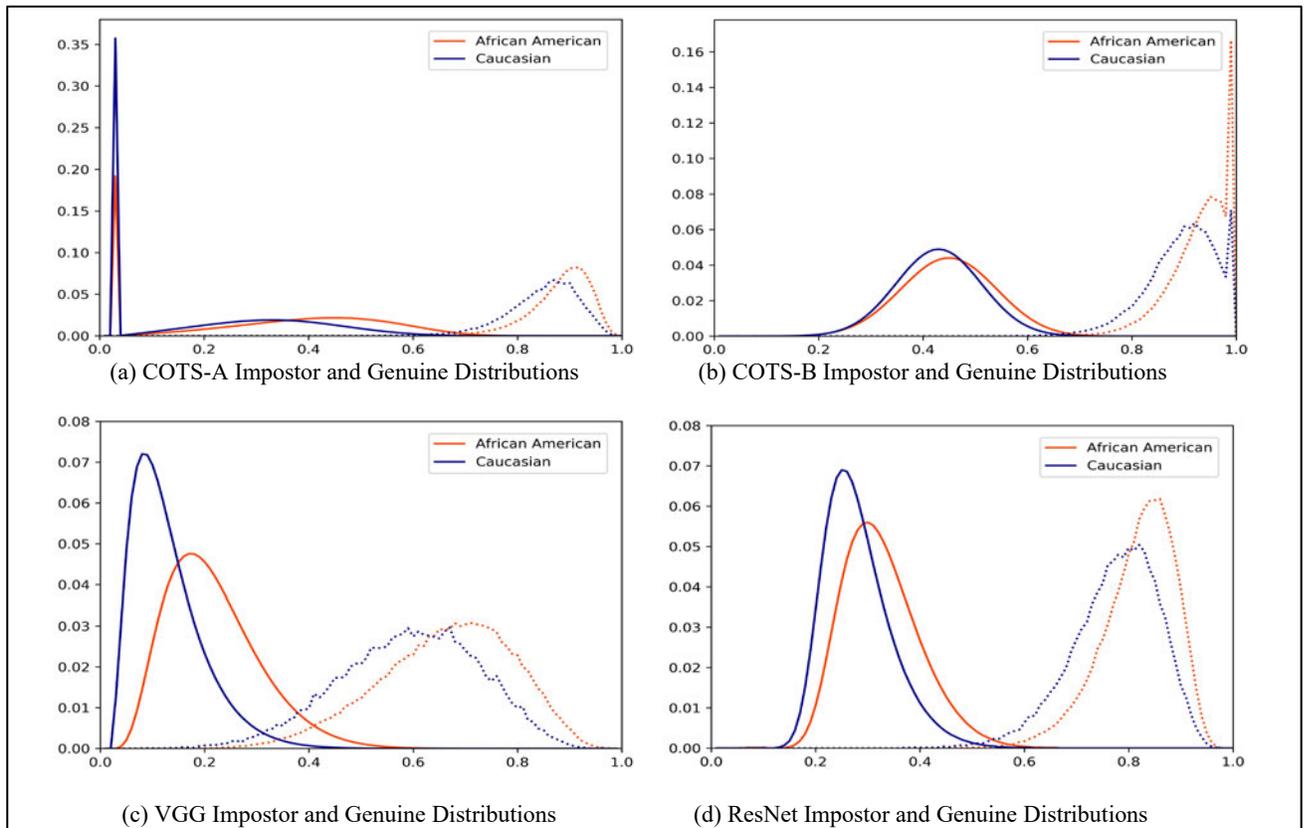

Figure 4 – Comparison of African-American and Caucasian Impostor (solid) and Genuine (dotted) Distributions for Different Matchers on the MORPH dataset. Horizontal axes show match scores normalized to between 0 and 1 for each matcher's min and max reported score; vertical axes show % of distribution.

## 5. Impostor / Genuine Distributions

Figure 4 shows the distributions of impostor and genuine scores for the African-American and Caucasian cohorts, computed using each of the matchers. The COTS matchers have more unusually-shaped distributions than the CNN matchers. COTS-A has a spike in the fraction of impostor pairs given its lowest reported score. COTS-B has a spike in the genuine distributions at COTS-B's highest reported score. Independent of differences between matchers, there are consistent differences between the African-American and the Caucasian distributions across the matchers.

For all the matchers, the impostor distribution for African-Americans is shifted, relative to that for Caucasians, toward higher similarity scores. Thus, for a given decision threshold, the African-American impostor distribution will generally have a higher FMR than the Caucasian impostor distribution.

For all the matchers, the genuine distribution for African-Americans is also shifted toward higher similarity scores. Thus, for a given threshold on the match score, the African-American genuine distribution will generally have a lower FNMR than the Caucasian distribution.

As may seem apparent in Figure 4, the differences between the distributions are statistically significant (2-sample K-S test, 2-sided, alpha = 0.05). The differences in the impostor and genuine distributions feed directly into differences in the FMR and FNMR curves and less directly into differences in the ROC curves.

The d-prime statistic can be employed to characterize the degree of separation between the means of the impostor and genuine distributions. The distributions are required to be Gaussian for the d-prime measure to be applicable. Visual inspection of the distributions in Figure 4 suggests that none are perfectly Gaussian, with the COTS matchers each having one bi-modal distribution. ResNet clearly has higher accuracy than VGG, and even though the ResNet distributions are not perfectly Gaussian, we compare the d-prime for the two cohorts to get an idea of the relative separation of the impostor and genuine distributions. The d-prime computed for the African-American distributions is 6.69, and for the Caucasian distributions is 6.59. This suggests that the ability of the ResNet face matcher to separate the mean of the imposter and genuine score distributions for the African-American and Caucasian image cohorts is comparable.

In a presentation on the NIST "FRVT Ongoing" work, Grother [20] has pointed out similar differences in accuracy between African-American and Caucasian image cohorts at the level of impostor and genuine distributions. The FRVT Ongoing results are based on different datasets and matchers from those used in this paper, so the observation of similar results suggests that the underlying phenomena are indeed general.

The inability to be conclusive in findings drawn from the ROC curve is mainly attributed to the absence of the decision threshold needed to achieve the specified FMR. While all of the information used to formulate the ROC is drawn from the impostor and genuine match score distributions, it is not common practice to note the decision threshold that generated the FMR-TPR pair. This is not particularly important when comparing accuracy of multiple algorithms on the same dataset, but is necessary when comparing accuracy of an algorithm on different cohorts.

## 6. Equal Image Quality = Equal Accuracy?

No previous study that compares recognition accuracy for African-American and Caucasian cohorts, or for that matter across any other demographic cohorts, has considered whether the quality of the images is similar between cohorts. Datasets such as MORPH [11,12] and the PCSO dataset [9] contain mugshot-style images. This implies a controlled acquisition, at defined locations, with a plain gray background behind the subject. For this reason, there may be an initial presumption that demographic splits of the dataset should result in approximately equal image quality. However, this is not the case, at least for the MORPH dataset as evaluated in terms of ICAO compliance by the tool used in this study.

A common standard of image quality relevant to face recognition is "ICAO compliance". The International Civil Aviation Organization has guidelines for face images used in travel documents, defined in ISO/IEC standard 19794-5 [17]. These guidelines have been implemented in automated form by a number of organizations; see [18] for a description of one well-known effort. We used the IFace SDK [19] to check images for ICAO compliance. Using default settings for the IFace ICAO compliance check, just over 48% of the African-American images were rated as ICAO compliant, and just over 57% of Caucasian images were rated as ICAO compliant. (For comparison, ICAO-compliance rates for male versus female images were nearly equal, within 1% for African-American images and within 2% for Caucasian images.) In particular, the scores for the "brightness" element of the ICAO compliance check appear to be distributed significantly differently for the African-American versus Caucasian cohorts. (We should note that it is possible that other tools for checking ICAO compliance may yield different results. The ICAO standard specifies aspects to assess for image quality, but the algorithm implemented to assess a particular aspect may vary.)

The fact that the African-American and Caucasian image sets differ in average image quality, as shown by the ICAO compliance rates, suggests the question – Does applying the same threshold for image quality to the image cohorts affect the apparent difference in accuracy?

Figure 5 shows ROC curves analogous to those in Figure 2, but computed only on the subset of images rated as ICAO-compliant. Note that the overall accuracy is greater than for the ROC curves in Figure 2. Also note that, for the matchers for which the ROC for the African-American cohort shows lower accuracy than the ROC for the

Caucasian cohort, the gap between the African-American and Caucasian ROCs is slightly reduced when using only the ICAO-compliant subset. For example, for the COTS-A matcher, the ROCs for the original image cohorts showed substantially better accuracy for the Caucasian cohort. But the COTS-A ROCs for the ICAO-compliant subset show a smaller gap, and nearly the same verification rate at a FMR of 1-in-100,000. The improvement is more marginal for the higher accuracy matchers.

## 7. Conclusions

The following conclusions can be summarized based on our experimental results to date.

Based on the face detection rates and the failure-to-enroll rates in Section 3, we find no good evidence for a difference in the face detection or failure-to-enroll rate between the African-American and Caucasian cohorts.

Using a known, available dataset resulting from an operational law enforcement scenario, and without applying any filter on image quality, we find that African-American image cohort is disadvantaged on FMR and advantaged on FNMR compared to the Caucasian image cohort.

Across a set of two COTS matchers and two well-known CNN matchers, we find that the differences in impostor and genuine distributions work out so that two matchers have a better ROC for the Caucasian cohort and two have a better ROC for the African-American cohort. (See Section 4). However, the more important point is that ROC curves are not an appropriate way to compare face recognition accuracy across demographic cohorts.

When ICAO compliance is used to select subsets of the images that are more equal on image quality, we find that the low-similarity tail of the genuine distribution is reduced for both cohorts.

Demographic balance was not a design goal the for VGGFace dataset [13], used to train VGG, or for the VGGFace2 dataset [15], used to train ResNet. Based on manual inspection of VGGFace2, we estimate the ratio of Caucasian to African-American subjects in the VGGFace2 dataset as in the range of 5:1 to 6:1. Yet the d-prime values for the ResNet impostor and genuine distributions show that inherent face recognition accuracy is at least as good for the African-American cohort as for the Caucasian cohort. ResNet's better ROC or higher d-prime for the African-American cohort was ***not*** achieved through a demographically-balanced training dataset, demonstrating that, at least in this instance, balanced training data is not a requirement to obtain balanced, or better, accuracy.

## 8. Discussion

*Why are your results are at odds with those of previous works?* Our results obtained using the COTS-A and VGG matchers broadly agree with previous works. Our results obtained with the COTS-B and ResNet matchers show that the pattern of ROC results seen in previous works is not general across all matchers. At least two factors may be involved. One is that COTS-B and ResNet are newer matchers than those used in previous works, and face recognition technology has improved over time. A second factor is that previous works did not identify accuracy differences at the level of the impostor and genuine distributions. If previous works had identified the pattern of the African-American cohort having higher FMR combined with lower FNMR, it may have suggested that the ROC for the African-American cohort could in principle be better or worse. More fundamentally, ROC curve comparisons are not an appropriate way to compare accuracy across demographic cohorts for an operational scenario that uses a fixed decision threshold.

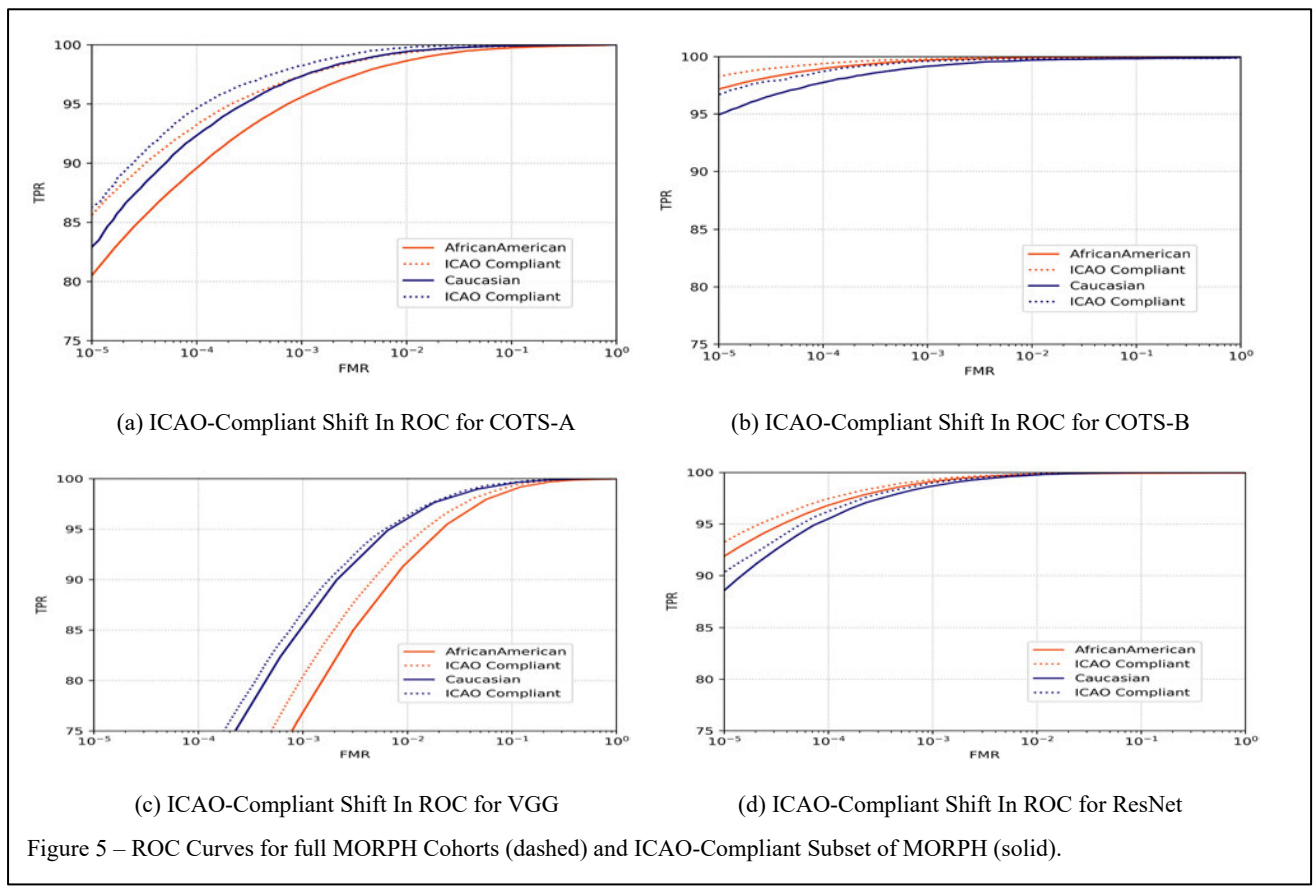

(a) ICAO-Compliant Shift In ROC for COTS-A

(b) ICAO-Compliant Shift In ROC for COTS-B

(c) ICAO-Compliant Shift In ROC for VGG

(d) ICAO-Compliant Shift In ROC for ResNet

Figure 5 – ROC Curves for full MORPH Cohorts (dashed) and ICAO-Compliant Subset of MORPH (solid).

*Why does the African-American image cohort have a lower rate of ICAO compliance?* Examination of the distributions of scores for elements of the ICAO-compliance check suggests that the distribution of image brightness scores is a big, perhaps the main, factor. At this time, we can only speculate on why the African-American image cohort has a larger fraction of poorly-lit images. One speculation is that the lighting should be adjusted according to the skin tone of the subject in order to obtain the best images for each person, and that subject-dependent lighting adjustment is currently generally not done at image acquisition.

*How can equal image quality be obtained at the time of image acquisition?* SDKs that automatically check for ICAO compliance are readily available. They can indicate specific issues, such as image too dark, subject not looking at camera, subject wearing glasses, etc. In scenarios where image acquisition is supervised, it should be possible to acquire an ICAO-compliant face image. Consider an analogy to the normal use of iris recognition. A commercial iris sensor like the Iris Guard AD 100 takes iris images continuously, checking focus and other quality metrics, providing subject positioning feedback, and possibly turning on a light to cause pupil constriction, in order to take an image that passes the quality checks. Face image acquisition with a check for ICAO-compliance, the possibility of adjusting lighting, and possible instructions to "look directly at the camera", would be an analogous acquisition process.

*Can ICAO-compliant images, as an example of "good quality" images, be obtained for "all" persons?* Most likely not; a person with heavy sideburns and beard and an eye patch may not easily be able to take an ICAO-compliant image. But in terms of skin tones, we speculate that lighting can be adjusted to allow ICAO-compliant images for all normal skin tones.